# TOPIC MINING BASED ON FINE-TUNING SENTENCE-BERT AND LDA

Jianheng Li [1] and Lirong Chen [2]

[1] Department of Computer Engineering, Inner Mongolia University, Hohhot , China
[2] School of Computer Science, Inner Mongolia University, Hohhot , China

**ABSTRACT**

*[Research background] With the continuous development of society, consumers pay more attention to the key information of product fine-grained attributes when shopping. [Research purposes] This study will fine tune the Sentence-BERT word embedding model and LDA model, mine the subject characteristics in online reviews of goods, and show consumers the details of various aspects of goods. [Research methods] First, the Sentence-BERT model was fine tuned in the field of e-commerce online reviews, and the online review text was converted into a word vector set with richer semantic information; Secondly, the vectorized word set is input into the LDA model for topic feature extraction; Finally, focus on the key functions of the product through keyword analysis under the theme. [Results] This study compared this model with other word embedding models and LDA models, and compared it with common topic extraction methods. The theme consistency of this model is 0.5 higher than that of other models, which improves the accuracy of theme extraction.*

## 1. INTRODUCTION

In the past, consumers mainly understood the quality of products by comparing store ratings, the number of product reviews, and product sales. Focus only on the overall aspects of the product, but ignore the various details of the product. Online comments have become an important way for people to express their personal opinions[1]. Consumers are more inclined to share their feelings and opinions about product usage on the platform. Digging useful information from online comments plays an important role in helping consumers understand various aspects of product information.

At present, commonly used text topic recognition techniques are divided into two categories: the first category is text topic recognition techniques based on traditional methods, and the second category is text topic recognition methods based on deep learning techniques. The traditional methods for text topic recognition include statistical based methods and rule-based methods. In statistical methods, this includes using one-hot encoding [2] to represent words, applying bag of words modelsto represent sentences, and using syntax based N-gram [3] language models for joint probability representation of sentences. Rule based topic recognition technology typically refers to machine learning based text topic recognition methods, including common clustering algorithms such as LDA topic models [4] and text clustering methods. Researchers constantly improve the traditional LDA model to continuously improve the accuracy of topic extraction.

BTM [5] (Biterm Topic Model, abbreviated as BTM) and WNTM (Word co-occurrence Network Topic Model) are two models with good performance. However, traditional topic extraction methods do not consider the relationship between contextual semantics of words, and only classify comment texts with high-frequency words as having the same topic, which still has shortcomings.

In order to obtain key topic information from a large number of online comments, Blei, David M, Ng, Andrew Y, Jordan et al. [6] began using traditional LDA (Latent Dirichlet Allocation) to mine topic information in online comments. Muhammad M et al. [7] collected feedback information from smart wearable devices and analyzed it using LDA topic extraction models to uncover hot topics that users are interested in at different times; Muntadher S et al. [8] collected review information from e-commerce platforms, analyzed product review information through LDA topic extraction models, and extracted user feedback on various aspects of the product, such as product quality, product packaging, and after-sales service.

The LDA topic model has a certain degree of simplicity and can extract topic features from documents. However, LDA topic models often rely on the assumptions of Dirichlet distribution and bag of words model when analyzing the topics of online comment texts, which assume that words are independent and ignore the semantic connections between words, and cannot fully reveal the potential interrelationships between topics in comment texts.

With the continuous deepening of deep learning research and the increasingly wide range of applications. Combining the advantages of LDA for global modeling of topics and the ability of Word2vec to capture local contextual lexical relationships, the LDA2vec model [9] was proposed. This model outperforms traditional LDA models in terms of performance. Reference combines LDA model and Bi LSTM model to analyze the topic and emotional information in online course comments. Although deep learning methods such as word2vec can effectively reproduce the contextual relationship of static words and improve the accuracy of word vectorization, they still cannot completely solve the problems of polysemy and limited semantic information. In addition, the continuous development of pre training language models has become a new direction of text topic extraction research.

Sentence-BERT is an optimization of BERT model. It is an optimization to improve the inaccurate representation of BERT in word vector. Improve the accuracy of subsequent tasks by improving the accuracy of word vectorization of comment text. Sentence-BERT is a language representation model based on pre training. It no longer uses the traditional one-way language model or shallow splicing of two one-way language models for pre training like the previous model, but embeds these words according to the semantic information of the text word context to achieve two-way representation. This method can better vectorize words and effectively solve the problem of polysemy. Fine tuning Sentence-BERT model in the field of online comments can make Sentence-BERT model have a more accurate understanding of unique words in the field of online comments, and improve the accuracy of subsequent topic extraction tasks.

In response to the above issues, this article utilizes a combination of fine-tuning the Sentence-BERT model[10] and LDA topic model to improve topic recognition performance. The model proposed in this article retains the advantages of traditional LDA topic models while enhancing the consistency of topics and the accuracy of subdividing categories.

## 2. MODEL CONSTRUCTION

This article proposes a method that combines the Sentence-BERT model and LDA model to fine tune the field of e-commerce online comments through the MLM method. The fine tuned

Sentence-BERT model converts online comment documents into word vectors and sends each document's word vector set to the LDA topic extraction model for topic mining. Generate sentence vectors for documents based on word vectors, combine online comment text sentence vectors with document topic vectors to generate T-SNE graphs[11] for text vector clustering, and observe the text clustering effects of different word embedding models.

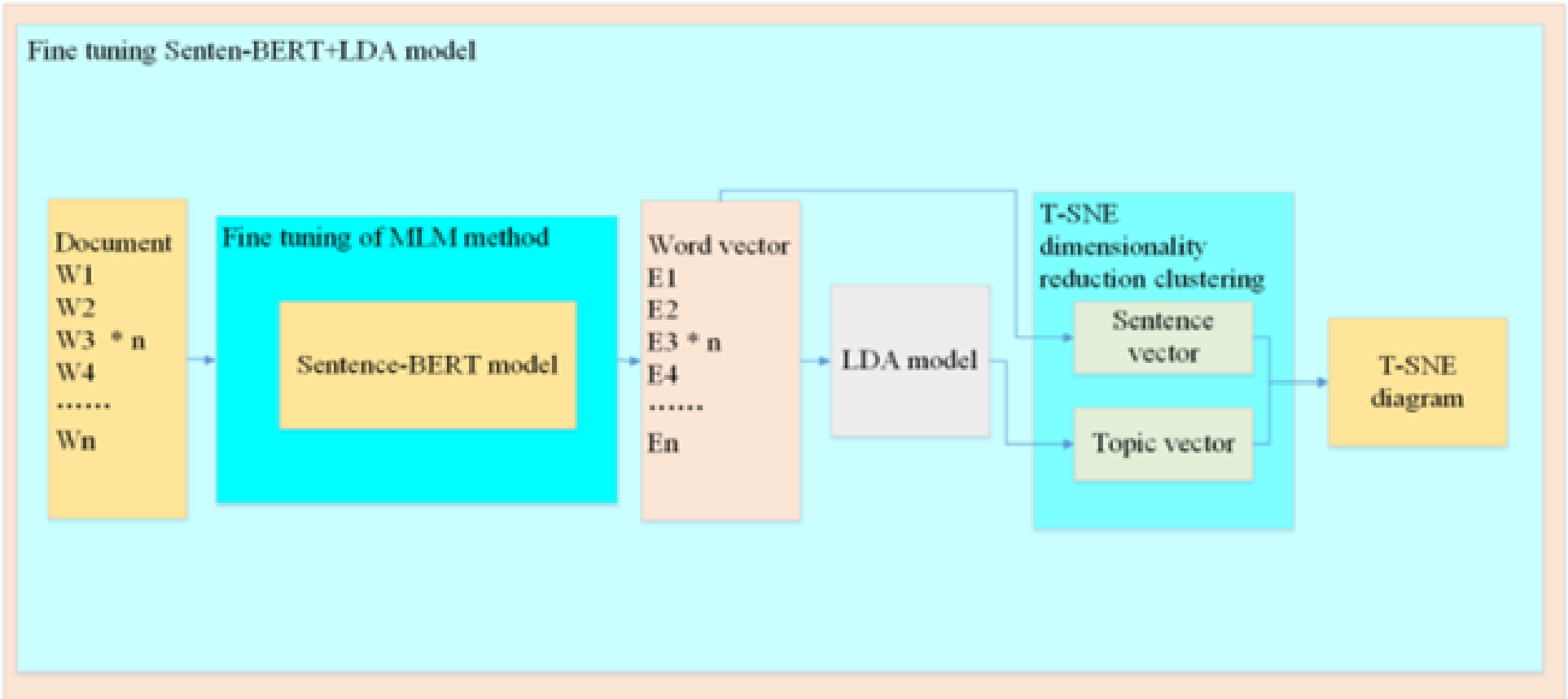

Figure 1. Fine tuning the Sentence-BERT+LDA model

## 2.1. Fine tune Sentence-BERT

Fine tune operation is a method of further training pre trained models to adapt to different domains. Sentence-BERT is a pre trained language model, however, in certain specific tasks, it may not fully capture the language patterns in the task. For example, there are differences in the way language is described in the automotive and e-commerce fields. Therefore, this article uses the Masked Language Model (MLM)[12] to fine tune the Sentence-BERT model. MLM method randomly shields words in the text, and then let the model predict these screened words to learn the context representation of words.

The specific operation of MLM method includes the following steps.

(1) Masking input words: During the training process, randomly select a portion of words from the input text and replace them with a special masking marker "[MASK]".
(2) Predicting obscured words: The model predicts these obscured words based on contextual information.
(3) Calculate loss: Use the cross entropy loss function to calculate the difference between the predicted words and the actual words in the model.
(4) Update model parameters: Use backpropagation algorithm to update model parameters and gradually optimize model performance.

MLM training formula:

Given an input sequence, where is the i-th word in the sequence. During the training process, randomly select a portion of words for masking:

$$X_{masked} = (\dots, [MASK], \dots) \quad (1)$$

The goal of the model is to predict obscured words based on context. Assuming the masked word is$x_i$, the model outputs$y=(y_1, y_2, y_3, \dots, y_n)$, $y_i$ is the probability distribution of the i-th word predicted by the model.The loss function is cross entropy loss:

$$\mathcal{L} = -\sum_{i\in masked} \log P(y_i = x_i | X_{masked}) \quad (2)$$

Among them, $P(y_i = x_i | X_{masked})$ represents the probability that the predicted word by the model is equal to the actual word.

## 2.2. T-SNE

T-SNE (T-Distributed Stochastic Neighbor Embedding) is a machine learning algorithm used for dimensionality reduction and data visualization. By analyzing the T-SNE visualization, we can observe and analyze the following aspects:

Topic clustering situation: Observe whether the points in the T-SNE graph exhibit obvious topic clustering structures. If the points of different themes can be clearly clustered together in the graph, and there are clear boundaries between different themes, it indicates that the theme extraction algorithm has successfully divided the document into different themes.

Topic distribution: Observe the distribution of topic points in the T-SNE graph. If the topic points are evenly distributed in the graph, it indicates that the importance of each topic is similar, and the topic extraction algorithm can balance the weights between different topics well.

Document distribution: Observe the distribution of document points in the T-SNE graph. If there is a good correspondence between document points and topic points, that is, documents on the same topic are clustered together in the graph, while document points on different topics are distributed more widely, it indicates that the topic extraction algorithm can correctly divide documents into corresponding topics.

## 2.3. Topic Confusion

Topic Confusion is an indicator in the field of natural language processing that measures the complexity of text topics. It can reflect the degree of confusion in the text theme. The lower the perplexity value, the more accurately the text is classified under each topic. The specific formula for calculating perplexity is as follows:

$$perplexity(D_{test}) = exp(-\frac{\sum log\ p(w_d)}{\sum_{d=1}^{M} Nd}) \quad (3)$$

Among them, M represents the number of all texts (the number of documents), and$w_d$represents the number of words in the d-th text.

$$p(w_d) = p(z|d) * p(w|z, r) \quad (4)$$

Among them, z represents the topic, w represents the document, and r is the text topic distribution learned based on the training set. Simply put, the logarithmic function numerator of Perplexity represents the negative likelihood estimate of the possibility of generating the entire document set (reflecting the strength of the training set parameter generation ability). Since the probability value range is [0,1], the numerator value of the logarithmic function is a positive value and shows a positive correlation with the text generation ability; The denominator is the number of words in the entire document collection.

### 2.4. Latent Dirichlet Allocation

Latent Dirichlet Allocation (LDA) is a statistical model used for text data analysis, which can help identify topics in large sets of documents. The LDA model assumes that a document is composed of multiple topics, each topic consisting of a specific set of words, and the frequency of these words appearing in the document can reflect the importance of the topics[13]. The LDA model structure is shown in Figure 2

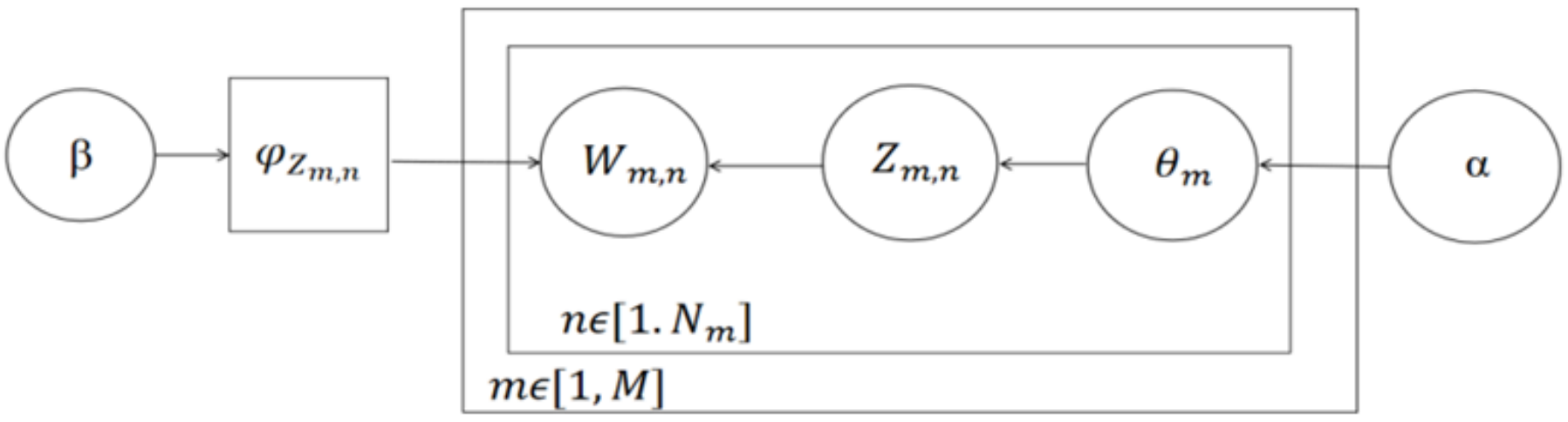

Figure 2. LDA model

$$perplexity(D_{test}) = exp\left(-\frac{\sum log\, p(w_d)}{\sum_{d=1}^{M} Nd}\right) \quad (5)$$

In Figure 2, $\theta_m$ represents the topic distribution of document m, a represents the $\theta_m$ prior distribution,$Z_{m,n}$represents the topic of the nth word sampled from document m, $\phi Z_{m,n}$represents the word distribution, β represents the prior distribution of the word distribution, $w_{m,n}$represents the nth word of the final generated mth article, Nm represents the total number of entries in document m, and there are a total of M documents.

$$p(w_{m,n}, z_{m,n}|a, \beta) = \sum_{n=1}^{N_m} p(w_{m,n}|z_{m,n}, \beta)p(z_{m,n}|a) \quad (6)$$

$p(w_{m,n}|z_{m,n})$ calculates the probability of each sampled term under the topic The probability distribution calculation formula for each entry in the mth document is as follows:

$$p(w_{m,n}) = \sum_{n=1}^{N} p(w_{m,n}|z_{m,n})p(z_{m,n}) \quad (7)$$

negative likelihood estimate of the likelihood of generating the entire document set (reflecting the strength of the training set parameter generation ability). Due to the probability range of [0,1], according to the definition of the logarithmic function, the numerator value is a positive value and positively correlated with the text generation ability; The denominator is the number of words in the entire document collection.

## 3. Experimental Process and Result Analysis

### 3.1.Data Collection and Preprocessing

To verify the effectiveness of the fine-tuning Sentence-BERT+LDA model proposed in this article, the experimental data used in this article is sourced from the e-commerce product review dataset previously published in previous articles, in order to validate the effectiveness of the model proposed in this article. The dataset contains online comment texts for different products. This article selects 30000 online comments for three categories: clothing, fruits, and mobile phones, to verify the effectiveness of the model proposed in this article.

In the field of natural language processing, text data preprocessing is an important task. This article uses a Chinese online comment dataset, and the data preprocessing is divided into the following parts: removing stop words and converting text data into vector format.

## 3.2. Evaluation

Due to the fact that most topic models are unsupervised methods, they do not have clear evaluation metrics such as accuracy and F1 score like supervised learning tasks. Therefore, topic consistency [26] is used to measure the quality of topic models, and the topic is scored by calculating the similarity between keywords under the topic.

$$TC(Z,W^2) = \sum_{t=2}^{T}\sum_{l=1}^{t-1} log_2 \frac{N(w_t^{(z)}, w_l^z) + \theta}{N(w_l^z)} \quad (8)$$

In the formula: $W^{(z)} = [w_1^z, w_2^z, w_3^z \ldots\ldots w_T^z]$is the set of top T keywords ranked in theme Z; $N(w_t)$represents the number of online comments with keywords $w_t$ appearing in the online comment collection; $N(w_t,w_l)$represents the number of online comments with words$w_t$ and $w_l$ appearing simultaneously in the online comment collection; θ is a constant, usually set to 1.

This article uses two methods, "u_mass" value and "c_v" value, to calculate the consistency of online comment topics. Among them, "u_mass" uses conditional probability to obtain the consistency score of the topic. By calculating the frequency of each binary appearing together in the set window range as a percentage of the total number of times it appears in the corpus, the frequency is logarithmically calculated, and the average value obtained is used as the representation of u_mass.

C_v is a co-occurrence based method for measuring topic consistency. It combines document co-occurrence statistics and word vector representation to measure topic consistency by evaluating the co-occurrence of topic words in the document. The calculation formula is as follows:

$$C_v = \frac{1}{|W|} \sum_{i=1}^{W} \sum_{j=i+1}^{W} NPMI(w_i, w_j) \quad (9)$$

Among them, W is a collection of theme words, $NPMI(w_i, w_j)$which is the standardized point mutual information between words$w_i$ and$w_j$.

The calculation of Normalized Pointwise Mutual Information (NPMI) is as follows:

$$NPMI(w_i, w_j) = \frac{PMI(w_i,w_j)}{-\log p(w_i,w_j)} \quad (10)$$

$$PMI(w_i, w_j) = \log \frac{p(w_i,w_j)}{p(w_i)p(w_j)} \quad (11)$$

## 3.3. Fine Tuning the Sentence-BERT Model Results

The fine-tuning operation data in this article is obtained by randomly selecting online comment data from the e-commerce dataset as the fine-tuning dataset. Firstly, randomly select 12000 online comment data as training data for the fine-tuning section; The loss variation of the Sentence-BERT model optimized by MLM method with training and testing sets allocated in an 8:2 ratio is shown in Figure 3.

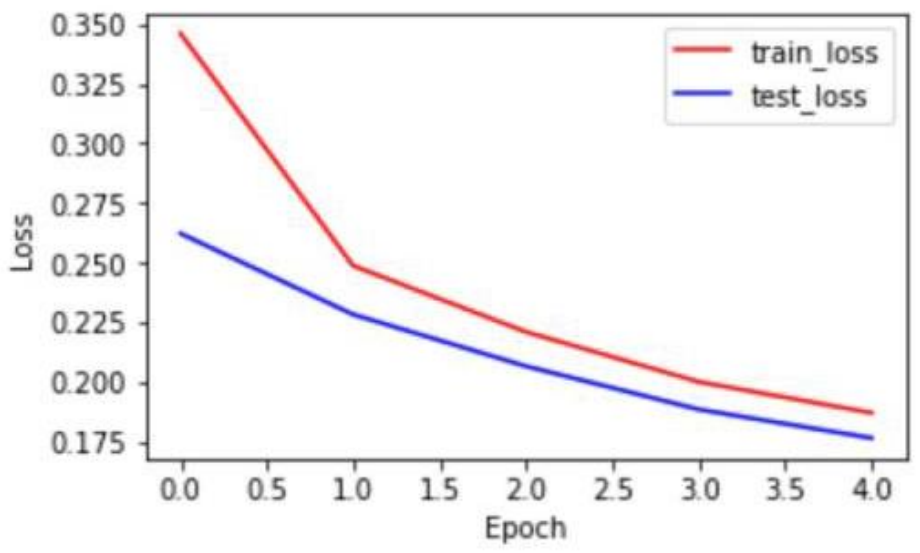

Figure 3. Loss Variation of MLM Method

## 3.4. Comparative Experimental Results and Analysis

This article first verifies the effectiveness of using the Sentence-BERT model for word vectorization. By comparing it with other word vectorization models, it proves the usefulness of using Sentence-BERT to generate word vectors for subsequent topic extraction. After verifying the effectiveness of the Sentence-BERT+LDA model for online comment topic extraction, the results of the Sentence-BERT+LDA model were compared with other excellent models in the field of text topic extraction to verify the superiority of our model.

To determine the optimal number of topics extracted by LDA[14] for different product reviews. This article uses the perplexity algorithm mentioned earlier to determine the optimal number of topics for different product datasets, and calculates the perplexity values for topics ranging from 1 to 14. The theme perplexity image of the fine-tuning Sentence-BERT[15]+LDA model is shown in Figure 4. In the comparative experiment, the theme consistency was obtained under the optimal number of themes, and the results are shown in Table1.

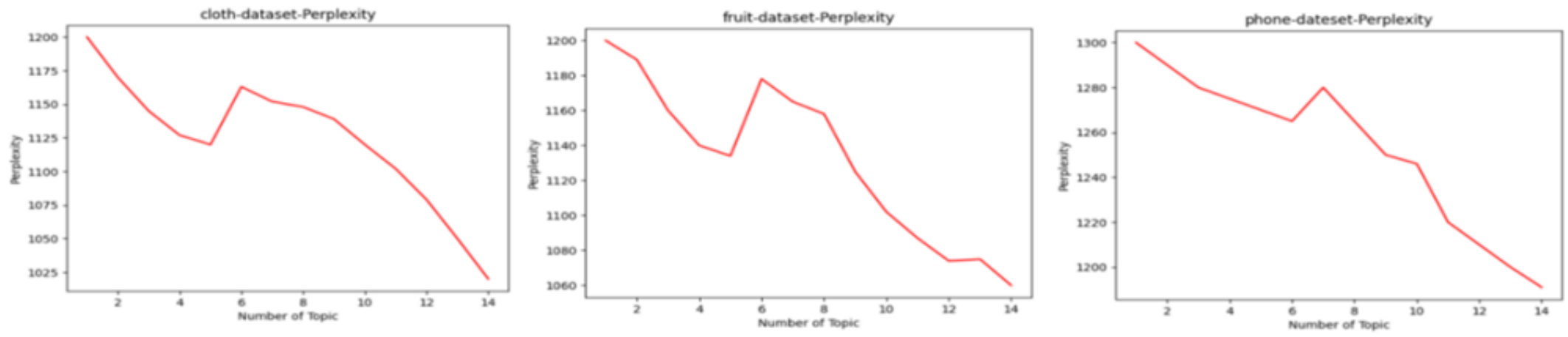

Figure 4. Sentence-BERT model+LDA perplexity graph for each dataset

Table 1. Experiment 1 Topic Consistency Table

| **Model** | **Clothing** | | **Fruit** | | **Mobile phone** | |
|---|---|---|---|---|---|---|
| | u_mass | c_v | u_mass | c_v | u_mass | c_v |
| TF-IDF+LDA | -5.56 | 0.55 | -5.65 | 0.54 | -5.62 | 0.45 |
| Word2Vec+LDA | -5.45 | 0.58 | -5.21 | 0.56 | -5.27 | 0.49 |
| BERT+LDA | -4.73 | 0.63 | -4.58 | 0.62 | -4.64 | 0.58 |
| Sentence-BERT+LDA | -4.25 | 0.67 | -4.39 | 0.65 | -4.54 | 0.62 |
| Model in this article | -4.17 | 0.69 | -4.35 | 0.68 | -4.23 | 0.64 |

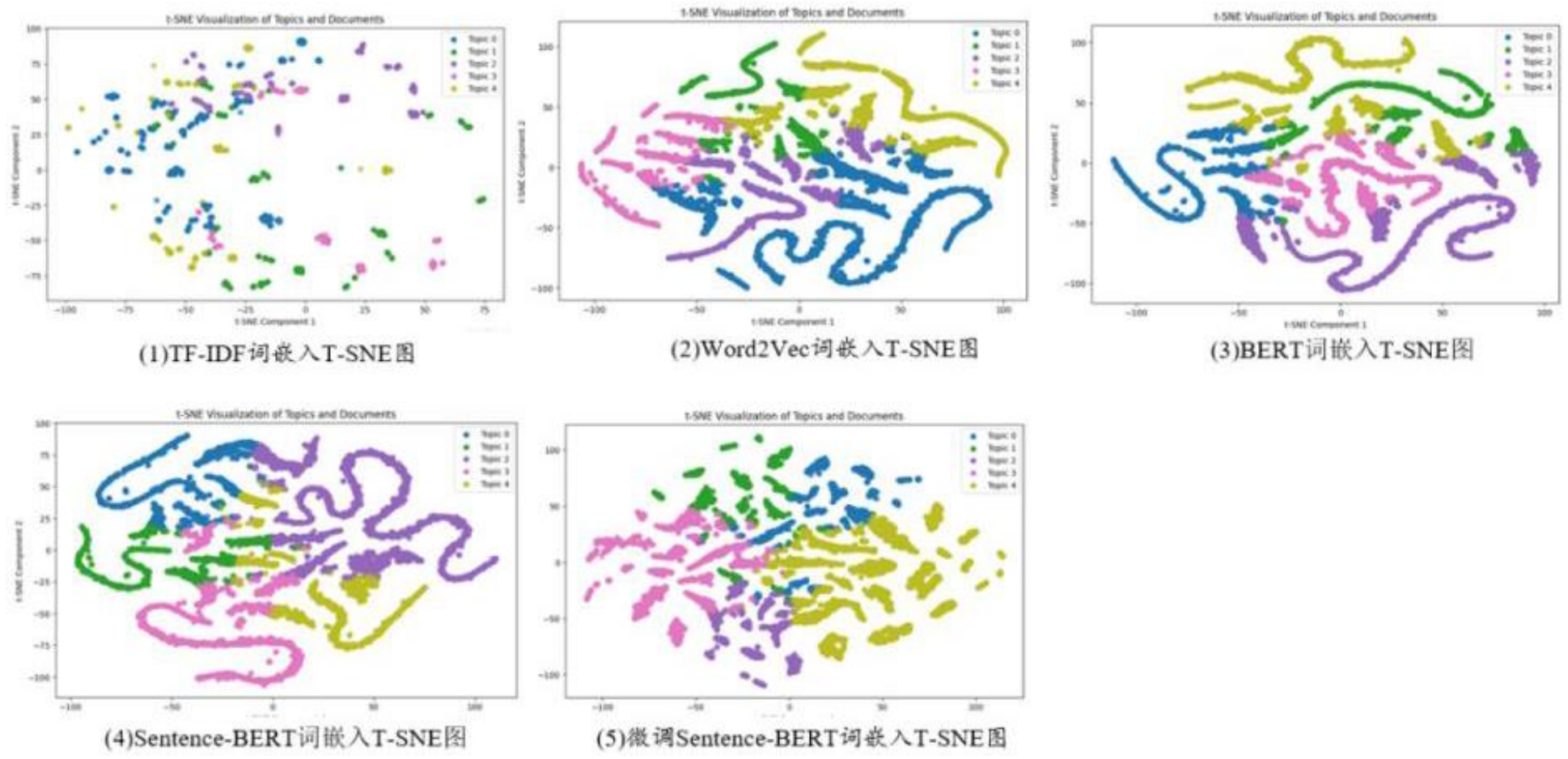

Figure 5. T-SNE diagram of different word embedding models for clothing data

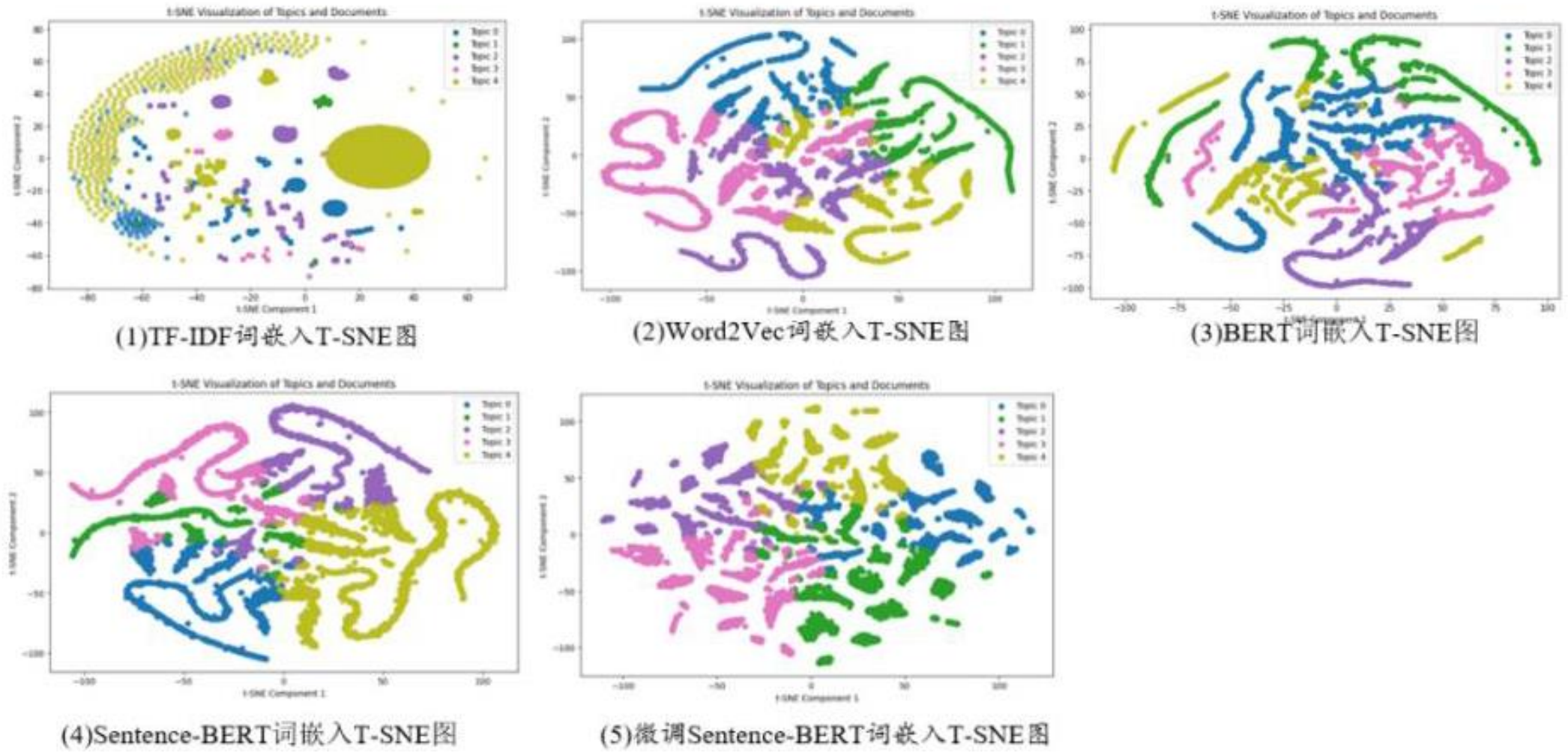

Figure 6. T-SNE diagram of different word embedding models for fruit data

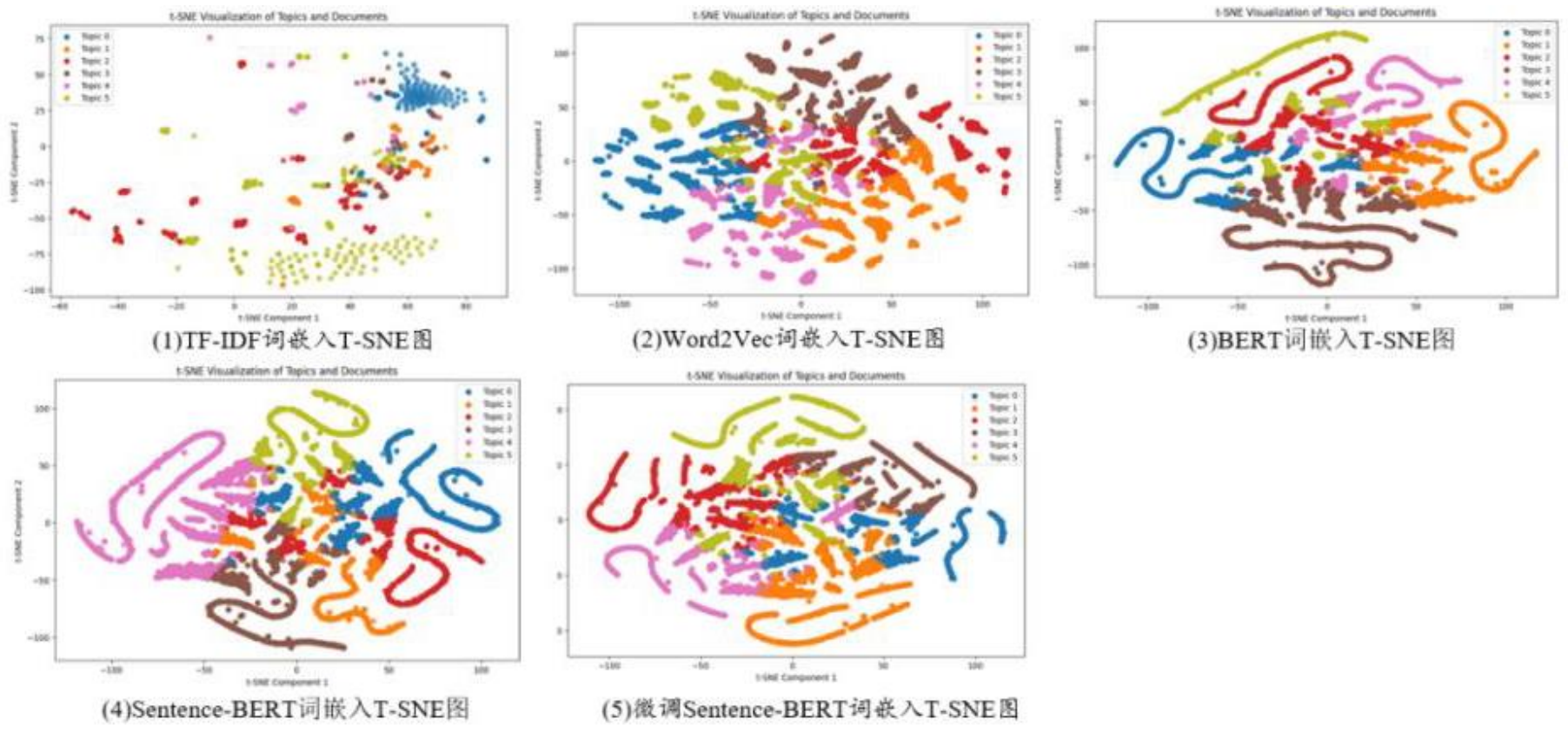

Figure 7. T-SNE diagram of different word embedding models for mobile data

Intuitively speaking, fine-tuning Sentence-BERT yields clusters that are relatively concentrated compared to other semantic feature extraction methods, and each cluster has a lower degree of overlap. Prove that fine-tuning Sentence-BERT has certain advantages.

Table 2. Fine tuning the Sentence-BERT+LDA model for topic extraction results

| **Data Set** | **Topic** | **Keyword** |
|---|---|---|
| Clothing | Topic1:Product Style | Style, pattern, genuine, fashionable, generous, pants shape, effect, pilling, product |
| | Topic2:Product quality | Fabric, size, quality, elasticity, thread ends, pure cotton, nylon, fuzzing, material, color difference |
| | Topic3:Product Logistics | Logistics, speed, shipping cost, physical item, suggestion, taste, exchange, pants legs, shipping cost |
| | Topic4:Product price | Price, affordable, not worth it, street vendor goods, discounts, cost-effectiveness, odor, price reduction |
| | Topic5:Product Service | Service attitude, attitude, customer service, exchange, counterfeit goods, enthusiasm, deception |
| Fruit | Topic1:Product taste | Taste, texture, too small, juicy, appearance, sweetness, appearance, refreshing |
| | Topic2:Product quality | Quality, quality, not worth it, bad fruit, attitude, experience, brand, brand |
| | Topic3:Product Logistics | Logistics, delivery, speed, cold chain, speed, situation, weight, box, refrigeration |
| | Topic4:Product price | Price, discount, affordability, special offer, flash sale, cost-effectiveness, quantity, deceiving people |
| | Topic5:Product Service | Service, Consumer, Trust, Attitude, Service Attitude, Customer Service, Influence, After sales |
| Mobile phone | Topic1:Product Performance | Resolution, system, screen, pixels, memory, performance, sound effects, calls, graphics |
| | Topic2:Product price | Price, gifts, cost-effectiveness, price reduction activities, price difference, affordable quality |
| | Topic3:Product accessories | Accessories: Headphones, Screen, Charger, Appearance, Protective Cover, Film Coating |
| | Topic4:Product quality | Quality, tone quality, product, quantity of electricity Crash, experience, brand, certified products |
| | Topic5:Product after-sales service | customer service, reply, After Sales, service Service attitude, passion, responsible, detection |
| | Topic6：Product Logistics | Logistics, speed, soon, not bad, send out goods deliver goods, receive, package, service |

Table 3. Experiment 2 Topic Consistency Table

| Model | Clothing | | Fruit | | Mobile phone | |
|---|---|---|---|---|---|---|
| | u_mass | c_v | u_mass | c_v | u_mass | c_v |
| BTM | -5.64 | 0.57 | -5.34 | 0.54 | -5.72 | 0.48 |
| LDA2vec[16] | -5.23 | 0.58 | -5.12 | 0.60 | -5.25 | 0.54 |
| BERTopic[17] | -4.65 | 0.62 | -4.74 | 0.63 | -4.82 | 0.62 |
| Model in this article | -4.17 | 0.69 | -4.35 | 0.68 | -4.23 | 0.64 |

From the analysis of the experimental results in Table 1, it can be seen that when using Sentence-BERT[18] on different product review data, the consistency value reaches its maximum value regardless of u_mass or c_v, proving that the Sentence-BERT model has certain advantages. TF-IDF[19] and Word2Vec cannot achieve good recognition of words with multiple meanings, namely polysemous words. Word2Vec relies on trained word vectors, each with a fixed word vector value, and does not determine the word vector based on actual contextual semantics. It is a static word vector mapping, which results in the generated word vectors being static and not changing with contextual changes, and may not be effective in handling dynamic contextual changes.

In the first comparative experiment, it has been verified that the Sentence-BERT word embedding model performs better than other word embedding models. In contrast experiment 2, the commonly used baseline models in the subject extraction field of the proposed model are compared to verify the effectiveness of the proposed model. Table 3 compares the results of Experiment 2, and it can be seen that compared with other topic extraction models, the topic consistency evaluation index of our model is the best. When models such as BERTopic[20] and LDA2vec extract feature information from online comment texts, they do not solve the problems of polysemy and multi word synonymy in the word vector representation of online comment texts, resulting in insignificant topic extraction effects.

## 4. CONCLUSION

This study will combine the fine-tuning of the Sentence-BERT word embedding model with the LDA topic model. Solved the problems encountered in LDA models, such as insufficient semantic description of comment texts, inability to effectively capture inter word correlations, and weak correlation of identified topic words. Analyze the advantages and disadvantages of various thematic features of the product. In the future, we will continue to explore how to balance the relationship between coarse and fine granularity in clustering processing, as well as how to use more reasonable fine-tuning strategies to further improve the effectiveness of topic extraction.

## REFERENCES

[1] Tan X .Chinese Internet Buzzwords: Research on Network Languages in Internet Group Communication[J].Information & Culture,2022,57(3):350-352.

[2] Liang Jie, Chen Jiahao, Zhang Xueqin, etc Anomaly detection based on single hot encoding and convolutional neural network [J]. Journal ofTsinghua University (Natural Science Edition), 2019, 59 (07): 523-529.

[3] Suyanto S ,Andi S ,Nafi R I , et al.Augmented-syllabification of n-gram tagger for Indonesian words and named-entities[J].Heliyon,2022,8(11):e11922-e11922.

[4] Jianhang Z ,Qi Z ,Shaoning Z , et al.Latent Linear Discriminant Analysis for feature extraction via Isometric Structural Learning[J].Pattern Recognition,2024,149110218-.

[5] Zhang F .Improved BTM topic embedding method for Web text data extraction[J].Entertainment Computing,2024,50100642-.

[6] Blei D M, Ng A Y, Jordan M I. Latent dirichlet allocation [J]. Journal of machine Learning research,2003,3(Jan):993-1022.

[7] Muhammad M ,Furqan R ,Fahad A , et al.What people think about fast food: opinions analysis and LDA modeling on fast food restaurants using unstructured tweets.[J].PeerJ. Computer science,2023,9e1193-e1193.

[8] M H H A ,Muntadher S ,Badrul N A .Web content topic modeling using LDA and HTML tags.[J].PeerJ. Computer science,2023,9e1459-e1459.

[9] Jianfeng X ,Yunhe Z ,Zhiqiang L , et al.A Recognition Method of Truck Drivers' Braking Patterns Based on FCM-LDA2vec[J].International Journal of Environmental Research and Public Health,2022,19(23):15959-15959.

[10] Naofumi F ,Yasuhiro O ,Shoji K .Accuracy of the Sentence-BERT Semantic Search System for a Japanese Database of Closed Medical Malpractice Claims[J].Applied Sciences,2023,13(6):4051-4051.

[11] Pouyet E ,Rohani N ,Katsaggelos K A , et al.Innovative data reduction and visualization strategy for hyperspectral imaging datasets using t-SNE approach[J].Pure and Applied Chemistry,2018,90(3):493-506.

[12] Mateusz B ,Masashi T ,Dusan R , et al.MIss RoBERTa WiLDe: Metaphor Identification Using Masked Language Model with Wiktionary Lexical Definitions[J].Applied Sciences,2022,12(4):2081-2081.

[13] Zhang J ,Gui W ,Wen J .China's policy similarity evaluation using LDA model: An experimental analysis in Hebei province[J].Journal of Information Science,2024,50(2):515-530.

[14] Zhao X ,Peng Z ,Fu S .Knowledge Graph of Low-Carbon Technologies in the Energy Sector and Cost Evolution Based on LDA2Vec: A Case Study in China[J].Sustainability,2024,16(17):7337-7337.

[15] Izumi M ,Jin'no K .Feature analysis of sentence vectors by an image-generation model using Sentence-BERT:Special Section on Recent Progress in Nonlinear Theory and Its Applications[J].Nonlinear Theory and Its Applications, IEICE,2023,14(2):508-519.

[16] Knowledge Graph of Low-Carbon Technologies in the Energy Sector and Cost Evolution Based on LDA2Vec: A Case Study in China[J].Sustainability,2024,16(17):7337-7337.

[17] Gou Y ,Chen Q .Discovery and Analysis of Key Core Technology Topics in Proton Exchange Membrane Fuel Cells Through the BERTopic Model[J].Energies,2024,17(21):5418-5418.

[18] Zhou H ,Huang W ,Li M , et al.Relation-Aware Entity Matching Using Sentence-BERT[J].Computers, Materials & Continua,2022,71(1):1581-1595.

[19] Zhou J ,Ye Z ,Zhang S , et al.Investigating response behavior through TF-IDF and Word2vec text analysis: A case study of PISA 2012 problem-solving process data[J].Heliyon,2024,10(16):e35945-e35945.

[20] Lu C ,Zhu L ,Xie Y , et al.Analysis of Hot Topics and Evolution of Research in World-class Agricultural Universities Based on BERTopic[J].Applied Mathematics and Nonlinear Sciences,2024,9(1).